\documentclass[11pt]{article}

\usepackage[a4paper,margin=1in]{geometry}
\usepackage{amsmath}
\usepackage{graphicx}
\usepackage{url}
\usepackage[T1]{fontenc}
\usepackage[hidelinks]{hyperref}
\usepackage{xcolor}
\usepackage{listings}
\usepackage{algorithm}
\usepackage{algpseudocode}
\usepackage{comment}

\title{
Quantum Genetic Optimization for Negative Selection Algorithms in
Anomaly Detection
}

\author{
Giancarlo P. Gamberi \\
Mackenzie Presbyterian University \\
\texttt{giangamberi@hotmail.com.br}
\and
Calebe P. Bianchini \\
Mackenzie Presbyterian University \\
\texttt{calebe.bianchini@mackenzie.br}
}

\date{}

\begin{document}

\maketitle

\begin{abstract}
Negative Selection Algorithms (NSAs), inspired by the self/non-self
discrimination mechanism of the human immune system, have been widely
employed in anomaly detection. However, their effectiveness is often
constrained by the efficiency of detector generation. This paper presents the
Quantum Genetic Negative Selection Algorithm (QGNSA), a novel approach that
integrates a Quantum Genetic Algorithm (QGA) into the EvoSeedRNSA algorithm,
replacing its classical evolutionary optimization process. The proposed method
exploits quantum superposition and probabilistic amplitude adjustment to
enhance search space exploration and convergence efficiency in the detector
generation process. Empirical evaluations using the Metaverse Financial
Transactions Dataset demonstrate that QGNSA achieves superior anomaly
detection accuracy compared to its classical counterpart while maintaining
robustness under varying hyperparameter configurations. The experimental
results highlight the potential advantages of quantum computing in artificial
immune systems, particularly in high-dimensional anomaly detection tasks.
Future research will focus on further optimizing quantum circuit design,
deploying the algorithm on real quantum hardware, and exploring hybrid
quantum-classical approaches for improved computational efficiency.
\end{abstract}


\section{Introduction}

Quantum phenomena have served as the foundation for the development of a wide
range of emerging technologies, such as quantum computing, quantum memories,
quantum teleportation, quantum key distribution, and quantum machine learning.
Among these, quantum computing exploits principles like superposition to enable
the processing of information in fundamentally different ways, offering
potential advantages over classical computing in solving certain computationally
hard problems~\cite{ZIDAN2023109844}.

The concept of utilizing quantum mechanics as the foundation for computation
began to take shape in the early 1980s, initially driven by the challenge of
efficiently simulating quantum systems using classical computers. It was
hypothesized that a quantum device, by operating under the same physical
principles as the systems being simulated, could offer a significant
computational advantage. Subsequent developments revealed that the potential of
quantum computing extends beyond quantum simulations, with promising
applications in solving a broad class of computational problems more efficiently
than classical approaches~\cite{EvolutionaryAlgorithmsAndQuantumComputing}.

Alongside the development of quantum algorithms, interdisciplinary efforts have
led to the integration of evolutionary computing principles with
quantum-inspired concepts, giving rise to Quantum-Inspired Evolutionary
Programming (QIEP). This approach has proven effective in tackling complex
optimization problems across a wide range of application domains~\cite{QIEP}.

Anomaly detection has greatly benefited from the application of various
artificial intelligence techniques, which, when augmented with human oversight,
contribute to enhanced system and data integrity. In parallel, advancements in
quantum computing have facilitated the execution of quantum algorithms on
commercially available, albeit still experimental, quantum hardware. Among these
are quantum machine learning algorithms, which hold the potential to leverage
quantum speed-up, thereby offering performance advantages over their classical
counterparts~\cite{CORLI2025107632}.

While reviewing the state-of-the-art in Negative Selection Algorithms
(NSA)~\cite{Gupta_2022}, a largely unexplored opportunity was identified:
leveraging quantum computing to enhance both performance and accuracy. Although
the concept of integrating quantum computing with negative selection algorithms
is not entirely new, limited exploration and the current lack of practical
implementations have highlighted a significant research gap. This gap served as
a key motivation for the development and investigation presented in this paper.

In this paper, we introduce the Quantum Genetic Negative Selection Algorithm
(QGNSA), a novel approach that draws inspiration from two existing
methodologies: the Quantum Genetic Algorithm (QGA) and the Evolutionary Seed
Real Negative Selection Algorithm (EvoSeedRNSA). By combining elements from
both, the proposed algorithm seeks to strengthen the performance of negative
selection through quantum-inspired optimization techniques, capitalizing on
quantum mechanical principles such as superposition and probabilistic evolution.

The structure of the paper is as follows:

\begin{itemize}
    \item Section~\ref{sec:related} reviews relevant literature on quantum
    negative selection algorithms and closely related works.

    \item Section~\ref{sec:quantum} outlines the key concepts of quantum
    computing, including qubits, superposition, entanglement, interference,
    and quantum noise, and provides a detailed overview of the Quantum
    Genetic Algorithm.

    \item Section~\ref{sec:nsa} introduces the core ideas behind negative
    selection algorithms and offers a comprehensive review of the
    EvoSeedRNSA model.

    \item Section~\ref{sec:qgnsa} presents our proposed Quantum Genetic
    Negative Selection Algorithm and compares it with its classical
    counterpart.

    \item Section~\ref{sec:experiments} describes the experimental setup and
    results, examining the algorithm performance under different parameter
    settings and benchmarking it against the classical EvoSeedRNSA.

    \item Section~\ref{sec:conclusion} concludes the paper and suggests
    avenues for future research.
\end{itemize}

\section{Related Works}\label{sec:related}

For this research, four academic databases were consulted to investigate the
extent of exploration into the combination of negative selection algorithms and
quantum computing. Although quantum evolutionary algorithms are gaining
attention, limited research specifically addresses their integration with
negative selection algorithms. All databases were last accessed on February 7,
2025.

Scopus was the first database explored. Using the Advanced Search feature, we
queried documents containing the terms ``quantum'' and ``negative selection'' in
the title, abstract, or keywords, along with ``algorithm'' in any part of the
document. This search yielded four results. Upon closer examination, the
documents, comprising two conference reviews~\cite{HIS2017,AMFSM2013}, one
journal article~\cite{Zhou2015499}, contained the specified keywords but did not
focus on quantum negative selection algorithms. The last document~\cite{QNSA}
proposes a quantum negative selection algorithm, which we will talk about later
in this section.

Next, we searched IEEE Xplore using the Advanced Search to find documents with
``quantum'', ``negative selection'', and ``algorithm'' in their metadata. This
search returned 48 documents, including~\cite{QNSA}. Although several promising
studies addressed quantum evolutionary algorithms (such as, but not limited
to~\cite{9606934,8817916,9996682,10712885}), none of them explored the
integration with negative selection algorithms.

The final two databases, ACM Digital Library and Web of Science, were queried
using similar criteria: the terms ``quantum'' and ``negative selection'' within
abstracts and keywords, and ``algorithm'' in general. In both cases, the search
yielded only the paper~\cite{QNSA}. These findings highlight a significant gap
in the literature, underscoring the novelty and importance of this research in
advancing the intersection of quantum computing and negative selection
algorithms.

In 2012, the work in~\cite{QNSA} was the first paper that described how to
integrate negative selection algorithms with quantum computing. Its main goal
was to address the associative algorithm process as an optimization problem,
aiming to identify optimal classification association rules (CAR) for
constructing an effective classifier. The proposed algorithm leveraged quantum
properties to generate and evolve populations of detectors through
quantum-inspired mutation mechanisms, iteratively refining the detector set
until the optimal CAR was identified. This pioneering approach demonstrated the
potential of quantum computing to improve the efficiency of negative selection
algorithms in classification tasks.

\section{Quantum Computing}\label{sec:quantum}

As advancements in anomaly detection continue to progress, the integration of
alternative computing paradigms has emerged as a pivotal research direction.
Although classical computational approaches have proven effective, they often
encounter limitations in efficiently managing complex, high-dimensional data.
Recent studies have investigated various methods to address these challenges,
including bio-inspired algorithms, neuromorphic computing, and quantum
computing. Among these approaches, quantum computing stands out for its
potential to substantially enhance optimization and machine learning tasks by
utilizing quantum superposition, entanglement, and interference.

Alongside neuromorphic computing, quantum computation and quantum simulation
have emerged as leading paradigms within the broader landscape of alternative
computing architectures. Among the most prominent implementations are platforms
based on superconducting qubits, which offer the potential to significantly
accelerate the solution of complex problems that are intractable or
computationally prohibitive for classical systems~\cite{CRANGANORE2024346}.

Over the past decade, quantum computing has progressed from a predominantly
theoretical framework to practical implementation, influencing a wide spectrum
of developments—including the design of algorithms and applications,
methodological advancements, and the integration of quantum components with
classical High-Performance Computing (HPC)
infrastructures~\cite{MARKIDIS2025107503}.

Due to its novel computational paradigm, quantum computing has emerged as a
rapidly expanding field of study, with ongoing research dedicated to exploring
its potential applications in areas such as cybersecurity and artificial
intelligence. This study focuses on a specific quantum algorithm, the Quantum
Genetic Algorithm~\cite{QGA}, which serves as a fundamental component in the
methodology of our proposed approach.

\subsection{Quantum genetic algorithm}

One of the fundamental quantum algorithms incorporated into our proposed
methodology is the QGA. As the name suggests, the QGA is a direct adaptation of
the Genetic Algorithm (GA), a bio-inspired heuristic designed to solve complex
optimization and search problems. The GA is fundamentally based on the
principles of natural selection and evolutionary biology.

The algorithm operates on a population of candidate solutions that evolve over
multiple generations through the application of selection, crossover, and
mutation operators. Initially, a random population is generated, with each
individual representing a potential solution to the given problem.

Each individual is then evaluated using a fitness function, which quantifies its
effectiveness in solving the problem. A selection process based on these fitness
values is conducted, favoring individuals with higher fitness scores for
reproduction. Various selection strategies can be employed, depending on the
specific implementation.

Next, the crossover operator is applied, where selected pairs of individuals
undergo genetic recombination, exchanging its composition to produce new
offspring. Following this, the mutation operator introduces random variations
into certain individuals, enhancing genetic diversity and preventing premature
convergence. This iterative process continues until a predefined stopping
criterion, such as a maximum number of generations or an optimal fitness
threshold, is reached.

Evolutionary computing methods that employ qubit-based representations exhibit
enhanced diversity compared to classical approaches, owing to their ability to
encode solutions in a superposition of states. This intrinsic property
facilitates broader exploration of the solution space. Additionally,
convergence behavior can be naturally modeled within the qubit paradigm: as the
probability amplitudes $|\alpha_i|^2$ or $|\beta_i|^2$ tend toward 1 or 0, the
qubit chromosome collapses to a definitive state, leading to a gradual loss of
diversity. Consequently, the qubit representation inherently supports both
exploration and exploitation, enabling a dynamic balance between these two
fundamental aspects of evolutionary optimization~\cite{QGA}.

In the classical genetic algorithm (GA), candidate solutions can be encoded
using various representations, including binary, numerical, and symbolic. In
contrast, the Quantum Genetic Algorithm (QGA) leverages qubits to represent
candidate solutions. This quantum encoding introduces a more diverse approach
due to the superposition property of qubits, allowing each individual to exist
in multiple states simultaneously. Additionally, convergence is achieved by
adjusting the probability amplitudes of qubit states, guiding the evolution
toward optimal solutions while preserving diversity through the inherent
randomness of quantum measurements. This characteristic enables the QGA to
explore the solution space more efficiently compared to its classical
counterpart.

The algorithm begins by initializing a quantum circuit $Q(t)$ in an equal
superposition state. The initial population $P(t)$ is generated by measuring
$Q(t)$, where the observed qubits represent the candidate solutions. Each
individual in the population may be encoded using one or more qubits, with a
higher qubit count providing greater precision in representation. The population
is then evaluated using a fitness function, which quantifies solution quality
based on the specific problem requirements, analogous to the classical approach.
The individual with the highest fitness value is retained as the best solution.

To update the quantum circuit, a set of quantum gates $U(t)$ is modified based
on the best solution found so far. Each gate performs a rotation operation along
the $Y$-axis, with the rotation angles adjusted to incrementally align with the
reference solution's state. A new quantum circuit $Q$ is then constructed, the
updated gates $U(T)$ are applied, and the circuit is measured, yielding a new
population $P(t)$. This iterative process continues until a predefined number
of generations is reached or an optimal fitness threshold is achieved. The
pseudo-code for the algorithm can be observed in Algorithm~\ref{alg:QGA}.

\begin{algorithm}[!ht]
\caption{Quantum Genetic Algorithm (QGA) -- adapted from~\cite{QGA}}
\label{alg:QGA}
\begin{algorithmic}[1]
    \Procedure{QGA}{}
        \State $t \gets 0$
        \State Initialize $Q(t)$
        \State Make $P(t)$ by observing $Q(t)$ states
        \State Evaluate $P(t)$
        \State Store the best solution among $P(t)$

        \While{not termination-condition}
            \State $t \gets t + 1$
            \State Make $P(t)$ by observing $Q(t - 1)$ states
            \State Evaluate $P(t)$
            \State Update $Q(t)$ using quantum gates $U(t)$
            \State Store the best solution among $P(t)$
        \EndWhile
    \EndProcedure
\end{algorithmic}
\end{algorithm}

\section{Negative Selection Algorithm}\label{sec:nsa}

The Negative Selection Algorithm (NSA) is a bio-inspired computational approach
particularly well-suited for tasks in which decision-making occurs within the
complementary space of a known positive profile. NSAs have demonstrated
effectiveness in various domains, including one-class classification, outlier
detection, and fault and intrusion detection. Drawing inspiration from the human
immune system, numerous NSA variants have been developed over the past three
decades, incorporating diverse representation schemes, distance metrics, and
coverage estimation strategies tailored to specific application
requirements~\cite{Gupta_2022}.

The NSA draws inspiration from the human immune system, mimicking the process by
which cells differentiate between self cells and foreign entities (non-self) to
combat potential threats. The algorithm functions by generating a set of
detectors, artificial entities that are classified as positive when they match
self data and negative when they match non-self data. These detectors are
subsequently utilized to classify new data into self or non-self categories.

Numerous variations of the NSA have been developed, each tailored to address
specific types of problems. This research focuses on the Real-valued Negative
Selection Algorithm (RNSA), which primarily employs derivatives of euclidean
distance for similarity matching. Specifically, we concentrate on the
EvoSeedRNSA algorithm, a variant that incorporates genetic algorithms to enhance
the generation of the detector set, thereby improving the accuracy in
distinguishing between self and non-self entities.

\subsection{EvoSeedRNSA Algorithm}

Our proposed negative selection variant is primarily based on the Evolutionary
Seed Real Negative Selection Algorithm. The EvoSeedRNSA algorithm, initially
introduced by~\cite{ZHANG1} and later extended to EvoSeedRNSAII
in~\cite{ZHANG2}, represents a significant advancement in negative selection
algorithms by integrating genetic algorithms to generate a more accurate
detector set. This algorithm forms the foundation for the quantum variant
developed in this research.

The original EvoSeedRNSA algorithm begins by generating a random population of
$N$ individuals, where each individual corresponds to a random seed vector with
a dimensionality equal to that of the problem space. The random seed of each
individual serves as the center for a candidate detector. If a candidate
detector does not match any of the self-samples, it is classified as a mature
detector. The algorithm continues generating detectors iteratively until a
predefined threshold or detector set size is achieved.

Once the detector set is generated, its effectiveness is evaluated using a
labeled test set. Let $TP$ represent the number of nonself samples correctly
identified by the detector set, and $TN$ denote the total number of nonself
samples present in the test set. The detection rate ($DR$) is then computed
using the following formula~\cite{ZHANG2}:

\begin{equation}
    DR = \frac{TP}{TN}
\end{equation}

The fitness of an individual is defined as the detection rate achieved by its
corresponding detector set~\cite{ZHANG2}.

Following the fitness evaluation, three genetic operators are sequentially
applied to evolve the population. The first operator is tournament selection,
wherein a predetermined number of individuals is randomly selected, and the
individual with the highest fitness is retained in an intermediate population.
Next, the crossover operator is applied, in which pairs of individuals from the
intermediate population are selected, and the seed vectors that comprise them
may be exchanged based on a defined crossover probability. Finally, the mutation
operator is employed, which selects random individuals from the intermediate
population and introduces new randomly generated seeds within those
individuals, subject to a specified mutation probability.

After applying these operators, the intermediate population is re-evaluated. If
the best individual from the original population exhibits a higher fitness value
than the best individual in the intermediate population, it replaces the
worst-performing individual in the latter. This iterative process continues
until the maximum number of generations ($MaxGen$) is reached. The original
pseudo-code is demonstrated in Algorithm~\ref{alg:EvoSeedRNSA}.

\begin{algorithm}[!ht]
\caption{Evolutionary Seed Real Negative Selection Algorithm adapted
from~\cite{ZHANG2}}
\label{alg:EvoSeedRNSA}
\begin{algorithmic}[1]
    \Require $MaxGen$ \Comment{Maximum number of generations}
    \Procedure{EvoSeedRNSA}{}
    \State $t \gets 0$ \Comment{Current generation number}
    \State Generate the initial population $\varepsilon_{0}$
    \State Evaluate the fitness value of each individual in $\varepsilon_{0}$

    \While{$t < MaxGen$}
        \State Selection
        \Comment{Generates intermediate population $\varepsilon_{t+S}$}
        \State Crossover
        \Comment{Generates intermediate population $\varepsilon_{t+S+C}$}
        \State Mutation
        \Comment{Generates intermediate population $\varepsilon_{t+S+C+M}$}
        \State Evaluate the fitness value of each individual in
        $\varepsilon_{t+S+C+M}$

        \If{fitness of best individual in $\varepsilon_{t}$ $>$ fitness of
        best individual in $\varepsilon_{t+S+C+M}$}
            \State Replace the worst individual in $\varepsilon_{t+S+C+M}$
            with the best from $\varepsilon_{t}$
        \EndIf

        \State $\varepsilon_{t+1} \gets \varepsilon_{t+S+C+M}$
        \State $t \gets t + 1$
        \Comment{A new population $\varepsilon_{t}$ is generated}
    \EndWhile

    \State Generate the detector set $D$ by the best individual in
    $\varepsilon_{MaxGen}$
    \State \Return $D$
    \EndProcedure
\end{algorithmic}
\end{algorithm}

\section{Quantum Genetic Negative Selection Algorithm}\label{sec:qgnsa}

After gaining a comprehensive understanding of the principles underlying
EvoSeedRNSA and quantum genetic algorithms, we identified a promising
opportunity to enhance the EvoSeedRNSA algorithm by replacing its classical
genetic algorithm with the quantum genetic algorithm. The objective of this
research is to improve the accuracy of detector generation, thereby enhancing
the overall effectiveness of anomaly detection.

However, this is not a straightforward substitution; due to the fundamental
differences between classical and quantum evolutionary processes, adaptations to
the algorithm are required. These modifications ensure that the quantum genetic
algorithm integrates effectively with the EvoSeedRNSA algorithm while leveraging
quantum properties such as superposition and probabilistic state manipulation to
optimize the detector generation process.

\subsection{Problem Definition}

The proposed algorithm aims to develop an artificial detector for a dataset
containing $N$ anomalies, each defined by $M$ features. The detector's
effectiveness is assessed based on its ability to accurately identify anomalies
while minimizing misclassification with normal data. This evaluation is
performed using the Euclidean distance between the artificial detector and each
sample, ensuring that only those within a predefined threshold are considered
matches.

Euclidean distance is chosen as the primary similarity measure due to its
effectiveness in direct comparison between the generated detector and anomalies.
The performance of the detector improves as it correctly identifies a higher
number of anomalies while minimizing both false positives and false negatives.
An optimal detector is one that maximizes anomaly detection accuracy while
maintaining a low error rate.

\subsection{Algorithm Steps}

\begin{figure}
    \centering
    \includegraphics[width=0.5\linewidth]{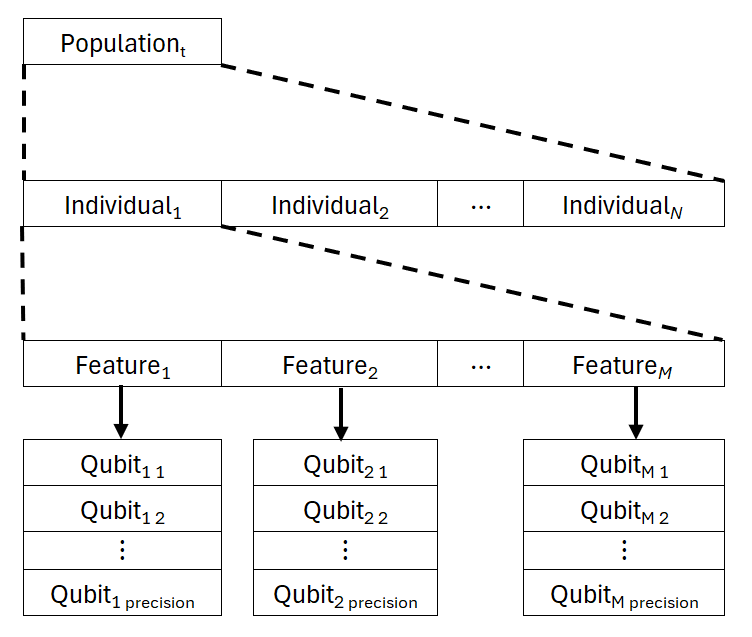}
    \caption{Diagram of the composition of a population ($t$) and of an
    individual.}
    \label{fig:Population-Diagram}
\end{figure}

For the QGA adaptation in our algorithm, each feature of the detector is
represented by a group of qubits, allowing for a more granular representation
of feature values. The individual detector corresponds to the measurement of the
entire quantum circuit, where the observed qubit states define the detector's
characteristics. The representation of this composition can be seen in
Figure~\ref{fig:Population-Diagram}.

To generate a population, multiple measurements, also called $shots$ by the
\text{qiskit} library, are made in the same quantum circuit. Each shot produces
a distinct detector, leveraging the inherent probabilistic nature of quantum
mechanics to explore a diverse range of potential solutions within a single
iteration.

The algorithm begins by initializing a quantum circuit with $M$ groups of
qubits, each group representing a feature of the generated detector. The
precision of each feature is determined by the size of these groups. Let the
group size be denoted as $precision$, resulting in a total of
$n = M \times precision$ qubits.

\begin{equation}
    n = M \times precision
\end{equation}

The $precision$ parameter determines the level of granularity in representing
each feature. A higher precision results in more accurate feature values, as
each feature is encoded using multiple qubits. For instance, setting precision
to 3 qubits allows each feature to assume $2^3$ distinct states. However,
increasing precision also leads to a proportional increase in the total number
of qubits required for the quantum circuit. Careful consideration is necessary
when selecting precision values to balance solution accuracy and computational
feasibility within available quantum resources.

An array $U$ is initialized with $n$ elements, each set to $\pi/2$ radians,
representing the initial angle $\theta$ for each qubit. These angles will be
adjusted throughout the algorithm to guide the evolution of the quantum state.

The algorithm iterates through generations, starting from $t = 0$ to an
arbitrary $MaxGen$, with the following process:

\begin{enumerate}
    \item The quantum circuit is measured repeatedly, with the number of
    measurements corresponding to the population size. Each measurement yields
    a temporary detector.

    \item The measurement results are translated into temporary detectors. Each
    measurement represents a unique detector configuration.

    \item The temporary detectors are evaluated by calculating their Euclidean
    distance to each anomaly in the training data set. The fitness value of a
    detector is determined by the proportion of anomalies it can detect, as
    defined by the threshold. If a temporary detector matches exactly an
    anomaly in the dataset, its fitness is set to zero to avoid duplication.

    \item If a temporary detector achieves a fitness value higher than the
    current best detector, it is designated as the new best detector. In the
    first generation, the first temporary detector serves as the initial best
    detector.

    \item If the fitness value of the best detector reaches $1$, indicating
    that it can detect all anomalies, the algorithm is terminated, returning
    this detector.

    \item For non-optimal fitness values:
    \begin{enumerate}
        \item If the fitness value of the best detector is 0, then no change is
        required.

        \item If the best detector qubit is measured as $1$, the corresponding
        $\theta$ in $U$ is increased by an adjustment coefficient $adj$, up to a
        maximum of $\pi$.

        \item If the qubit is $0$, $\theta$ is subtracted by $adj$, with a
        minimum of $0$.
    \end{enumerate}

    \item Y-rotations are applied to each qubit in the circuit, using their
    respective $\theta$ values from $U$.
\end{enumerate}

Both the convergence and exploration properties of the quantum algorithm are
evident in steps six and seven. Initially, the algorithm starts in an equal
superposition, resulting in a highly exploratory behavior, as the population
generation is entirely random. However, with each generation, the qubit
rotations are adjusted toward the best individual, increasing the likelihood of
measuring similar solutions. Despite this, the superposition of states ensures
that some degree of exploration is maintained, preventing premature convergence
and promoting diversity in the generated solutions. The algorithm steps are
outlined in Algorithm~\ref{alg:QGNSA}.

\begin{algorithm}[!ht]
\caption{QGNSA}
\label{alg:QGNSA}
\begin{algorithmic}[1]
    \Require $detectors\_train$, $populationSize$, $MaxGen$, $precision$, $evaluationThreshold$, $adj$
    \State $t \gets 0$
    \State $M \gets detectors\_train.\text{number\_of\_features}$
    \State $n \gets M \times precision$
    \State $\theta \gets \frac{\pi}{2}$
    \State $U \gets [\theta] \times n$
    \State $Best\_candidate \gets 0$
    \State Create a quantum circuit $QC$ with $n$ qubits
    \State initialize $QC$ in an equal superposition
    \While{$t < MaxGen$}
        \State Generate $Temporary\_detectors$ by measuring $QC$ $populationSize$ times
        \ForAll{$temporary\_detector \in Temporary\_detectors$}
            \State $Fitness \gets \text{evaluate}(temporary\_detector)$
            \If{$Fitness > Best\_candidate.\text{fitness}$}
                \State $Best\_candidate \gets temporary\_detector$
            \EndIf
        \EndFor

        \If{$Best\_candidate.\text{fitness} == 1$}
            \State \Return $Best\_candidate$
        \ElsIf{$Best\_candidate.\text{fitness} == 0$}
            \State Reset $QC$ and initialize it in an equal superposition
        \Else
            \State Adjust $U$ with the $Best\_candidate$
            \State Reset $QC$ and apply $Y\text{-}Gates$ with the angles of $U$
        \EndIf

        \State $t \gets t + 1$
    \EndWhile
\end{algorithmic}
\end{algorithm}

\subsubsection{Complexity analysis}

Asymptotically, the worst-case time complexity of the proposed algorithm, where
an optimal detector is either not found or is only identified in the final
generation, is given by $O(MaxGen*(PopulationSize*TrainSet+n))$, where $MaxGen$
denotes the predefined maximum number of generations,\break $PopulationSize$
represents the population size, $TrainSet$ corresponds to the number of
detectors in the training set, and $n$ is the circuit size, previously defined
as $n=M\times precision$. The two most computationally expensive components of
the algorithm are:

\begin{itemize}
    \item Fitness Evaluation: Each detector must be evaluated against the
    training set, which can be optimized by selecting only the best-performing
    detectors for training. This strategy maintains training efficacy while
    reducing the computational overhead of verification.

    \item Quantum Circuit Adjustment ($U$ Update): The angles of all quantum
    gates must be updated according to the best detector found in each
    generation. This step could be optimized by determining an optimal precision
    value, which balances solution space exploration and computational
    efficiency.
\end{itemize}

In the best-case scenario, where an optimal detector is identified in the first
generation, the asymptotic lower bound of the algorithm's complexity is given by
$\Omega(PopulationSize\times TrainSet+n)$. This is due to the necessity of
evaluating the initial population against the training set and performing the
quantum circuit measurement, but without the iterative process across multiple
generations.

For instance, the asymptotic complexity of the EvoSeedRNSA algorithm is
$O(MaxGen*PopulationSize*TrainSet)$. Although this may initially appear more
efficient compared to the quantum version, the actual runtime in real-world
scenarios can be significantly impacted by additional computational overhead.
As highlighted in~\cite{EvoSeedRNSA_TimeComp}, the size of the training set has
the most substantial influence on the performance of evolutionary negative
selection algorithms. This exacerbates the overhead introduced by the Selection,
Crossover, and Mutation operators, potentially offsetting the perceived
efficiency advantage of the classical approach.

\section{Experiments}\label{sec:experiments}

To assess the performance and effectiveness of our proposed algorithm, we
implemented both the classical and quantum versions of the EvoSeedRNSA algorithm
using Python. Furthermore, we incorporated functions from the \texttt{pandas}
and \texttt{numpy} libraries to facilitate data preprocessing, manipulation, and
transformation, ensuring efficient handling of the data set throughout the
experimentation process.

The classical version closely follows the original EvoSeedRNSA algorithm, with
minimal modifications. These modifications were primarily focused on adapting
the crossover and mutation functions to align with Python's syntax and utilize
the standard random library effectively. These changes ensured execution within
the Python environment while preserving the algorithm's functionality.

For the quantum implementation, the \text{Qiskit} library was used to simulate
the quantum components of the algorithm. The simulation necessitated a highly
capable solution to effectively handle the complexity and scale of the quantum
circuit, along with the requirement for executing multiple consecutive
measurements.

To meet these requirements, the \texttt{AerSimulator} was utilized,
specifically employing the \texttt{matrix\_product\_state} method. This method
was selected due to its efficiency and capability in handling the complexities
and subsequent measurements of our simulation.

\subsection{The anomaly dataset}

For the evaluation of the algorithms, we utilized the Metaverse Financial
Transactions Dataset from Kaggle~\cite{MetaverseDatabase}. As described: ``This
dataset provides blockchain financial transactions within the Open Metaverse,
offering a comprehensive, diverse, and realistic corpus for the development and
evaluation of anomaly detection models, fraud analysis, and predictive analytics
in virtual environments. It encompasses a wide array of transaction types, user
behaviors, and risk profiles across a global network''. This dataset aligns
precisely with the intended application of the EvoSeedRNSA algorithm.

The dataset comprises 78,600 transaction records, each containing 14 features.
Data preprocessing was conducted to prepare it for the algorithm distance
comparison. Initially, records were sorted by timestamp, which was subsequently
dropped. Columns such as IP prefix, sending address, receiving address, risk
score, and transaction type were also removed. These features were excluded due
to their limited relevance in detecting anomalies within the EvoSeedRNSA
algorithm or because they inherently indicate fraudulent transactions, which
could bias the detection process.

Normalization was then applied to the features for transaction amount, login
frequency, session duration, and hour of the day to standardize their scales.
For categorical features, including location region, purchase pattern, and age
group, one-hot encoding was performed to convert them into a numerical format
suitable for algorithmic processing. The anomaly feature was utilized to
categorize the dataset into two groups: a ``\text{mature detectors}'' group
comprising high-risk transactions, and a ``\text{self}'' group containing low
and moderate-risk transactions. Following this categorization, the anomaly
feature was removed from both groups to ensure unbiased training of the
algorithm. After the data preprocessing, the self set was reduced to 72,105
rows, while the mature detectors set contained 6,495 rows. Both sets retained
12 relevant features for further analysis.

To compare the accuracy of the algorithms, we employed a K-Fold cross-validation
strategy. The mature detectors group was partitioned into five folds, with each
Fold serving as a mature detector set and generating an artificial detector.
Given the stochastic nature of both algorithms, it was necessary to repeat each
fold multiple times to obtain a reliable estimate of their mean accuracy. For
this study, we arbitrarily selected five repetitions per fold, resulting in a
total of 25 runs for each algorithm. This approach provided a more comprehensive
assessment of the overall accuracy. Since the quantum version is being
simulated rather than executed on actual quantum hardware, execution time will
not be compared, as it would not provide a fair assessment.

The full dataset also underwent the same fold treatment. Each fold was utilized
to test the accuracy of the artificial detectors generated by both the classical
and quantum versions of the algorithm.

\subsection{Results}

\begin{figure}[!ht]
    \centering
    \includegraphics[width=0.5\linewidth]{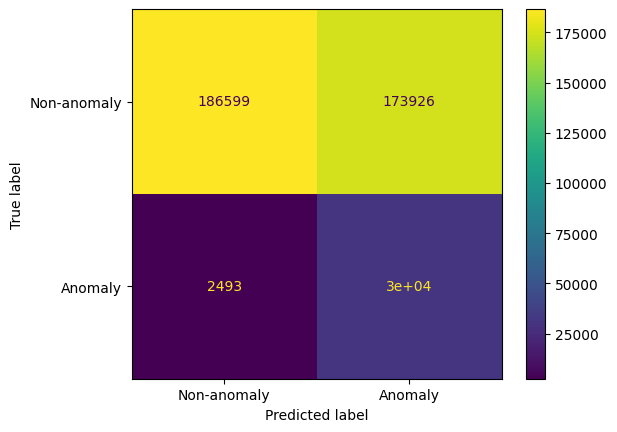}
    \caption{Confusion matrix for the first Quantum Algorithm execution.}
    \label{fig:QuantumConfMatrix1}
\end{figure}

\begin{figure}[!ht]
    \centering
    \includegraphics[width=0.5\linewidth]{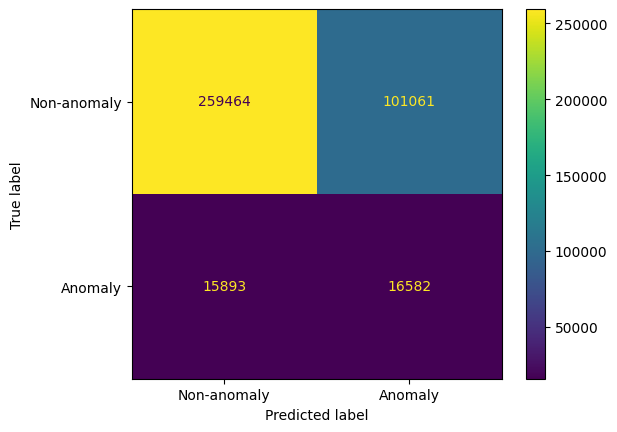}
    \caption{Confusion matrix for the Classical algorithm execution.}
    \label{fig:ClassicalConfMatrix}
\end{figure}

Following the execution of our algorithms within the test data, we recorded the
confusion matrices for all 25 iterations of each algorithm, as observed in
Figure~\ref{fig:QuantumConfMatrix1} and Figure~\ref{fig:ClassicalConfMatrix}.

Based on these matrices, we computed several key evaluation metrics to assess
the performance of the proposed approach. These metrics include:

\begin{itemize}
    \item \textbf{False Positive Rate (Type~I Error)}: This metric represents
    the proportion of negative samples that were incorrectly classified as
    positive. It is calculated as:
\end{itemize}

\begin{equation}
    False\ Positive\ Rate =
    \frac{False\ Positives(FP)}
         {False\ Positives(FP)+True\ Negatives(TN)}
\end{equation}

\indent A lower false positive rate indicates a reduced number of incorrect
anomaly detections.

\begin{itemize}
    \item \textbf{False Negative Rate (Type~II Error)}: This measures the
    proportion of actual positive samples that were misclassified as negative.
    It is given by:

\begin{equation}
    False\ Negative\ Rate =
    \frac{False\ Negatives(FN)}
         {False\ Negatives(FN)+True\ Positives(TP)}
\end{equation}

A lower false negative rate is crucial in anomaly detection, as it
minimizes missed detections.

    \item \textbf{Accuracy}: This metric quantifies the proportion of correctly
    classified instances among all instances:

    \begin{equation}
        Accuracy = \frac{TP+TN}{TP+TN+FP+FN}
    \end{equation}

    While accuracy provides a general measure of performance, it can be
    misleading in imbalanced datasets.

    \item \textbf{Precision}: Also known as positive predictive value,
    precision measures the proportion of correctly classified positive samples
    among all predicted positives:

    \begin{equation}
        Precision = \frac{TP}{TP+FP}
    \end{equation}

    Higher precision indicates fewer false positives, which is critical when
    false alarms need to be minimized.

    \item \textbf{Recall (Sensitivity)}: Recall represents the proportion of
    actual positives that were correctly classified:

    \begin{equation}
        Recall = \frac{TP}{TP+FN}
    \end{equation}

    A high recall is particularly important in anomaly detection, ensuring that
    most anomalies are correctly identified.

    \item \textbf{F1-Score}: The F1-score is the harmonic mean of precision and
    recall, balancing the trade-off between false positives and false
    negatives:

    \begin{equation}
        F1-Score = \frac{2*Recall*Precision}{Precision+Recall}
    \end{equation}

    A higher F1-score indicates a well-balanced model, particularly useful in
    cases where class imbalance is present.

    \item \textbf{Specificity}: This metric measures the proportion of actual
    negatives that were correctly classified:

    \begin{equation}
        Specificity = \frac{TN}{TN+FP}
    \end{equation}

    High specificity indicates that normal instances are less likely to be
    misclassified as anomalies.
\end{itemize}

Both algorithms were evaluated using identical parameters: 10 generations, a
population size of 10, and a threshold value of 1.6, which we considered to be
relatively stringent. For the quantum algorithm, a precision parameter of 16 was
employed. In contrast, the classical algorithm utilized a crossover probability
of 0.6 and a mutation probability of 0.4, values that are relatively high,
reflecting the need for a fair comparison against the quantum approach.

\begin{figure}[!ht]
    \centering
    \includegraphics[width=0.7\linewidth]{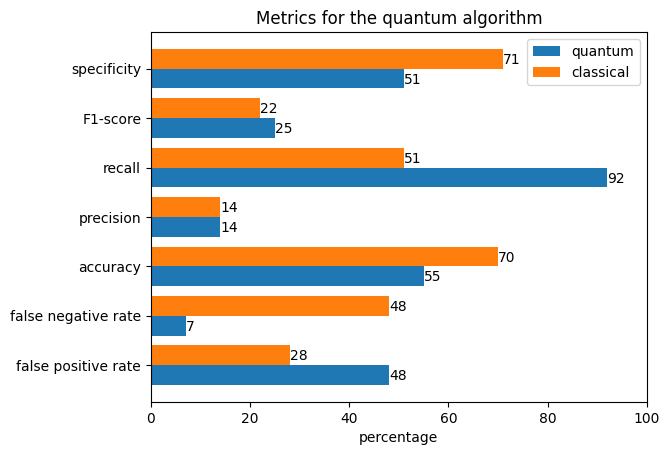}
    \caption{Metrics of the first Quantum algorithm and classical algorithm Metrics of the first Quantum algorithm executions.}
    \label{fig:CompMetrics}
\end{figure}

Upon analyzing the results, the most notable observation is that the quantum
version of the algorithm excelled in recall and had a lower false negative rate
(Type~II error), as seen in Figure~\ref{fig:CompMetrics}. This indicates that
it is highly effective at detecting anomalies, successfully identifying a larger
proportion of actual anomalies within the dataset. However, this improved
anomaly detection came at the cost of an increased false positive rate
(Type~I error), meaning that the quantum algorithm tends to misclassify normal
instances as anomalies more frequently.

In contrast, the classical algorithm demonstrated a more conservative approach,
yielding higher specificity and overall accuracy, as observed in
Figure~\ref{fig:CompMetrics}. This suggests that it is better at correctly
identifying normal instances, reducing false alarms. However, this cautious
classification comes with a drawback: the classical algorithm struggles to
detect anomalies effectively, leading to a higher number of false negatives.

\begin{figure}[!ht]
    \centering
    \includegraphics[width=0.5\linewidth]{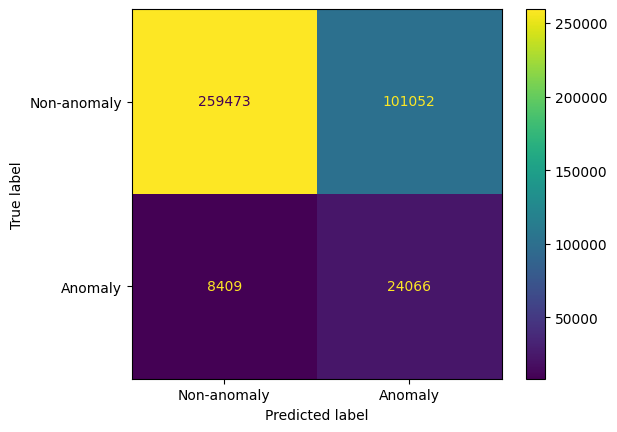}
    \caption{Confusion matrix of the second quantum test.}
    \label{fig:2ConfMatrix}
\end{figure}

\begin{figure}[!ht]
    \centering
    \includegraphics[width=0.5\linewidth]{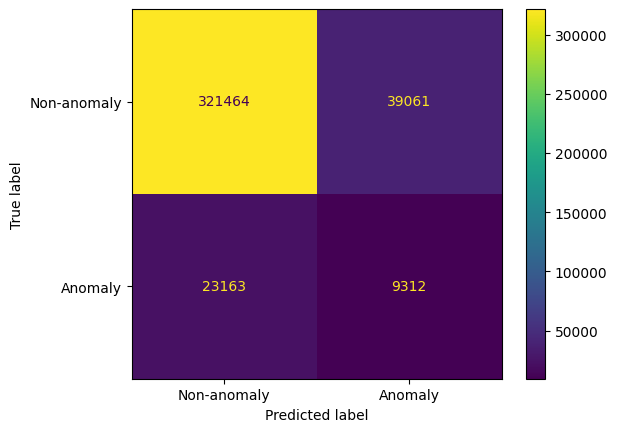}
    \caption{Confusion matrix of the third quantum test.}
    \label{fig:3ConfMatrix}
\end{figure}

\begin{figure}[!ht]
    \centering
    \includegraphics[width=0.5\linewidth]{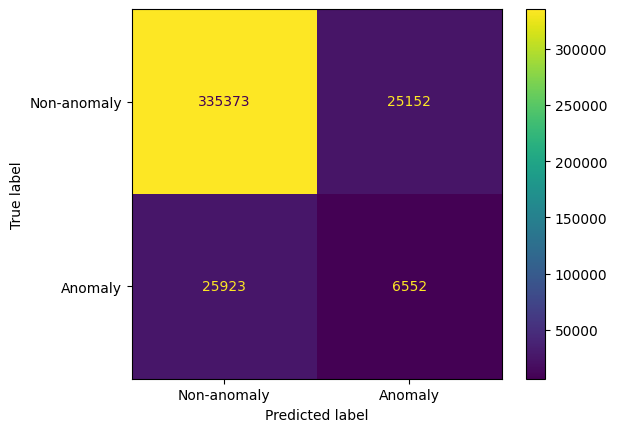}
    \caption{Confusion matrix of the fourth quantum test.}
    \label{fig:4ConfMatrix}
\end{figure}

\begin{figure}[!ht]
    \centering
    \includegraphics[width=0.5\linewidth]{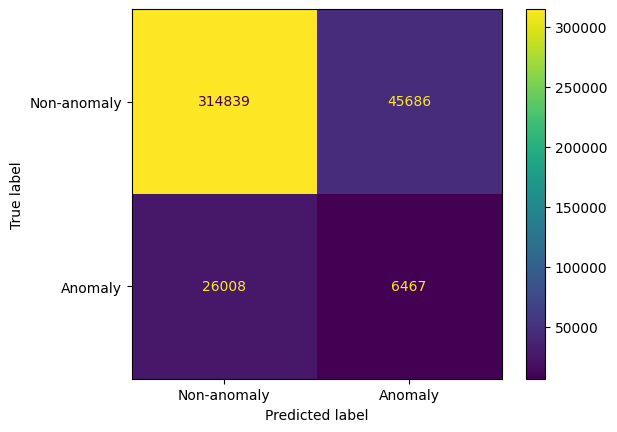}
    \caption{Confusion matrix of the second classical test.}
    \label{fig:2ConfMatrixC}
\end{figure}

\begin{figure}[!ht]
    \centering
    \includegraphics[width=0.5\linewidth]{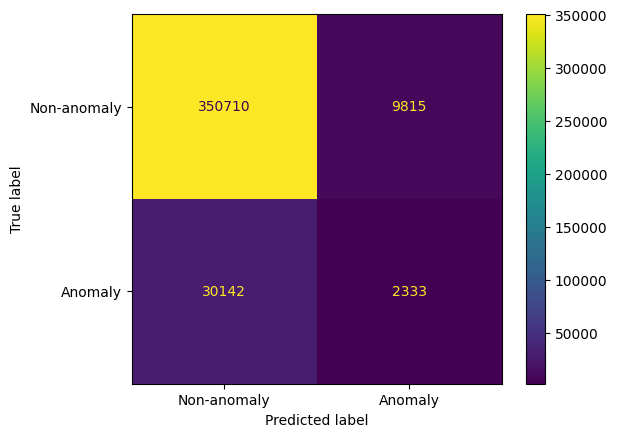}
    \caption{Confusion matrix of the third classical test.}
    \label{fig:3ConfMatrixC}
\end{figure}

\begin{figure}[!ht]
    \centering
    \includegraphics[width=0.5\linewidth]{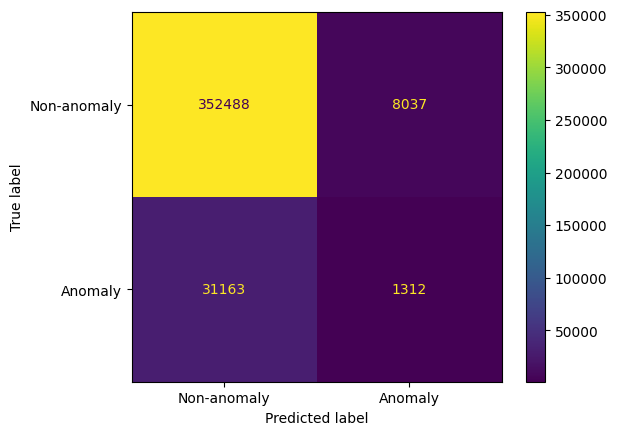}
    \caption{Confusion matrix of the fourth classical test.}
    \label{fig:4ConfMatrixC}
\end{figure}

To further fine-tune the quantum algorithm parameters and improve its
performance metrics, additional tests were conducted:

\begin{itemize}
    \item \textbf{Second Test}: The threshold was reduced to 1.4, and the
    precision was adjusted to 8. For the classical counterpart, only the
    threshold was modified. The corresponding confusion matrix can be observed
    in Figure~\ref{fig:2ConfMatrix} for the quantum algorithm and
    Figure~\ref{fig:2ConfMatrixC} for the classical.

    \item \textbf{Third Test}: The threshold was further reduced to 1.2,
    increasing the algorithm's strictness in detecting anomalies. The confusion
    matrix for this test is presented in Figure~\ref{fig:3ConfMatrix} for the
    quantum algorithm and Figure~\ref{fig:3ConfMatrixC} for the classical.

    \item \textbf{Fourth Test}: The $MaxGen$ and population size were both
    reduced to 5, and the precision was set to 4. In the classical counterpart,
    only the first two parameters were adjusted. The corresponding confusion
    matrix can be seen in Figure~\ref{fig:4ConfMatrix} for the quantum
    algorithm and Figure~\ref{fig:4ConfMatrixC} for the classical.
\end{itemize}

These parameter adjustments aimed to explore the impact of threshold reduction,
precision adjustments, and population constraints on the quantum algorithm's
performance, particularly in balancing anomaly detection effectiveness against
false positives.

\begin{figure}[!ht]
    \centering
    \includegraphics[width=0.7\linewidth]{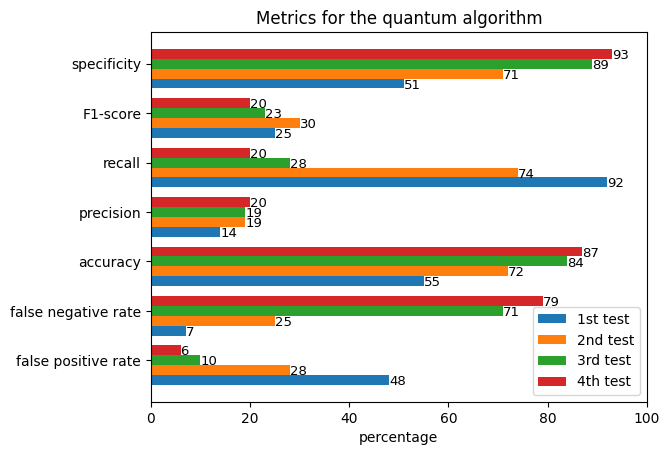}
    \caption{Metrics of the four tests of the quantum algorithm.}
    \label{fig:MetricsTests}
\end{figure}

\begin{figure}[!ht]
    \centering
    \includegraphics[width=0.7\linewidth]{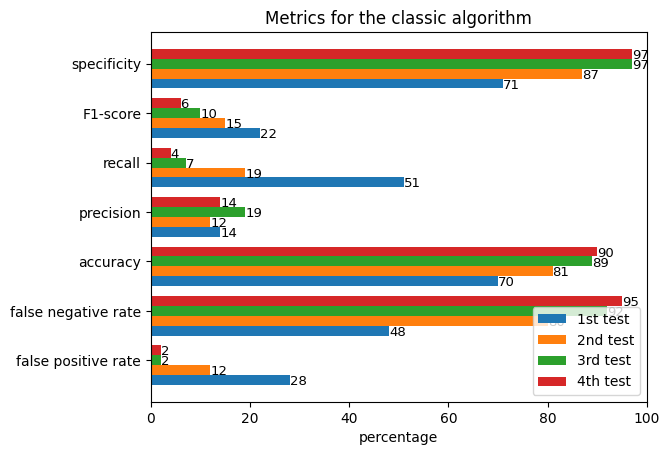}
    \caption{Metrics of the four tests of the classical algorithm.}
    \label{fig:MetricsTestsC}
\end{figure}

The resulting metrics of the quantum algorithm are compared in
Figure~\ref{fig:MetricsTests}, and in Figure~\ref{fig:MetricsTestsC} for the
classical.

Notably, lowering the threshold made the quantum algorithm highly specific and
more accurate, indicating that it became more conservative in classifying
anomalies. However, this came at the cost of reduced recall and an increased
false negative rate, meaning that while the algorithm was more confident in its
classifications, it missed more anomalies. A similar trend was observed in the
classical algorithm, though the impact on recall and the false negative rate
was more pronounced, suggesting a greater susceptibility to errors under
stricter threshold conditions.

On the other hand, reducing the number of generations, population size, and
precision did not significantly degrade the algorithm's performance. The most
significantly impacted metrics were the F1-score and the false negative rate,
highlighting a decline in the algorithms' ability to detect anomalies.

\section{Conclusion}\label{sec:conclusion}

The proposed quantum algorithm has demonstrated effectiveness in generating
artificial detectors capable of identifying anomalies while maintaining a high
level of consistency across various test scenarios.

Its adaptability makes it useful in domains with different anomaly detection
priorities. For example, in cybersecurity, high recall is crucial, even if it
leads to a higher false positive rate, as missing a real threat can have severe
consequences. In medicine, higher specificity is preferred, reducing false
positives to avoid unnecessary interventions while ensuring accurate anomaly
detection. This flexibility highlights the potential of the quantum approach in
tailoring detection strategies based on the specific needs of different fields.

Future research should focus on validating the algorithm across a diverse set of
datasets, particularly in domains where reliable anomaly detection is crucial,
such as cybersecurity for intrusion detection and healthcare for disease
diagnosis. Moreover, deploying the quantum algorithm on actual quantum hardware
is a critical next step. However, this presents a considerable challenge, as the
algorithm's qubit requirements can be substantial, contingent on the dataset
size and complexity.

For instance, our test configuration, which includes 12 features and a
precision of 16, necessitates quantum hardware with 192 qubits. As of late
2024, some of the largest available quantum computers include Atom Computing's
system with 1,180 qubits, IBM's Condor with 1,121 qubits, and Google's Willow
with 105 qubits. Consequently, our test could only be executed on the first two,
both of which are known to exhibit significant decoherence effects. However,
since our quantum circuits are relatively shallow, the impact of quantum noise
would be minimized, making execution more feasible.

The limited availability of quantum devices with the necessary qubit capacity
poses a significant hurdle, yet it is essential for evaluating the practical
viability of the proposed quantum approach.

Future research directions include leveraging quantum computing not only for the
evolutionary process but also for the evaluation phase, or even transitioning
the entire algorithm to a fully quantum implementation, thereby minimizing
reliance on classical computations.

\section*{Acknowledgments}

The authors express their appreciation for the foundational research that
informed and inspired this work. In particular, prior contributions in the
area of quantum computing and negative selection laid the groundwork for the
ideas developed in this study.

The authors thank MackCloud~(\url{https://mackcloud.mackenzie.br/}), the
Multidisciplinary Lab. for Scientific and Cloud Computing at Mackenzie
Presbyterian University, for their support. Grant~\#2018/25225-9, Sao Paulo
Research Foundation (FAPESP). Mackenzie Research and Innovation Fund
(MackPesquisa) -- Grants~\#231009 and~\#251005.

\bibliographystyle{unsrt}
\bibliography{ref}

\end{document}